\def\year{2022}\relax
\documentclass[letterpaper]{article} 
\usepackage{aaai22}  
\usepackage{times}  
\usepackage{helvet}  
\usepackage{courier}  
\usepackage[hyphens]{url}  
\usepackage{graphicx} 
\usepackage{natbib}  
\usepackage{caption} 
\DeclareCaptionStyle{ruled}{labelfont=normalfont,labelsep=colon,strut=off} 
\usepackage{algorithm}
\usepackage{algorithmic}
\usepackage{multirow}
\usepackage{graphicx}
\usepackage{subfigure}
\usepackage{amsfonts,amssymb}
\usepackage{newfloat}
\usepackage{listings}
\floatstyle{ruled}
\newfloat{listing}{tb}{lst}{}
\floatname{listing}{Listing}

\title{

FCA: Learning a 3D Full-coverage Vehicle Camouflage for Multi-view Physical Adversarial Attack
}
\author{
	Donghua Wang\textsuperscript{\rm 1}\equalcontrib,
	Tingsong Jiang\textsuperscript{\rm 2}\equalcontrib,
	Jialiang Sun\textsuperscript{\rm 2},
	Weien Zhou\textsuperscript{\rm 2},
	Xiaoya Zhang\textsuperscript{\rm 2},
	Zhiqiang Gong\textsuperscript{\rm 2},
	Wen Yao\textsuperscript{\rm 2},
	Xiaoqian Chen\textsuperscript{\rm 2}\thanks{Corresponding author.}
}
\affiliations{
	\textsuperscript{\rm 1}College of Computer Science and Technology, Zhejiang University\\
	
	\textsuperscript{\rm 2}
	Defense Innovation Institute, Chinese Academy of Military Science 

	wangdonghua@zju.edu.cn, tingsong@pku.edu.cn, sun1903676706@163.com, weienzhou@outlook.com, wendy0782@126.com, 
	\{gongzhiqiang13, zhangxiaoya09, chenxiaoqian\}@nudt.edu.cn
}

\usepackage{bibentry}

\begin{document}

\maketitle

\begin{abstract}
Physical adversarial attacks in object detection have attracted increasing attention. However, most previous works focus on hiding the objects from the detector by generating an individual adversarial patch, which only covers the \textit{planar} part of the vehicle’s surface and fails to attack the detector in physical scenarios for multi-view, long-distance and partially occluded objects. To bridge the gap between digital attacks and physical attacks, we exploit the \textit{full} 3D vehicle surface to propose a robust Full-coverage Camouflage Attack (FCA) to fool detectors. Specifically, we first try rendering the non-planar camouflage texture over the full vehicle surface. To mimic the real-world environment conditions, we then introduce a transformation function to transfer the rendered camouflaged vehicle into a photo-realistic scenario. Finally, we design an efficient loss function 
to optimize the camouflage texture. 
Experiments show that the full-coverage camouflage attack can not only outperform state-of-the-art methods
under various test cases but also generalize to different environments, vehicles, and object detectors. 
The code of FCA will be available at: \textit{https://idrl-lab.github.io/Full-coverage-camouflage-adversarial-attack/}.
\end{abstract}

\section{Introduction}
Over the past years, deep neural networks (DNNs) have achieved tremendous success in computer vision tasks.
However, DNNs are found vulnerable to adversarial examples \cite{szegedy2013intriguing}, which are elaborately designed to mislead DNNs to make incorrect predictions. As a new security issue in artificial intelligence, adversarial attacks appeal the attraction from both academics and industry. 

Adversarial attacks can be divided into two categories by their applicable domains: 1) \textbf{digital attacks} directly add imperceptible perturbations to pixels of input images in the digital space \cite{szegedy2013intriguing}, while 2) \textbf{physical attacks} modify objects in the real-world environment or physical simulators \cite{chen2018shapeshifter,sharif2016accessorize,kurakin2016adversarial,lu2017adversarial,athalye2018synthesizing} to investigate whether the perturbations are physically realizable and can stay adversarial under different transformations.
In this paper, we mainly concentrate on the latter as it is a more direct threat to visual systems in the physical world.

\begin{figure}
	\centering
	\includegraphics[width=1\linewidth]{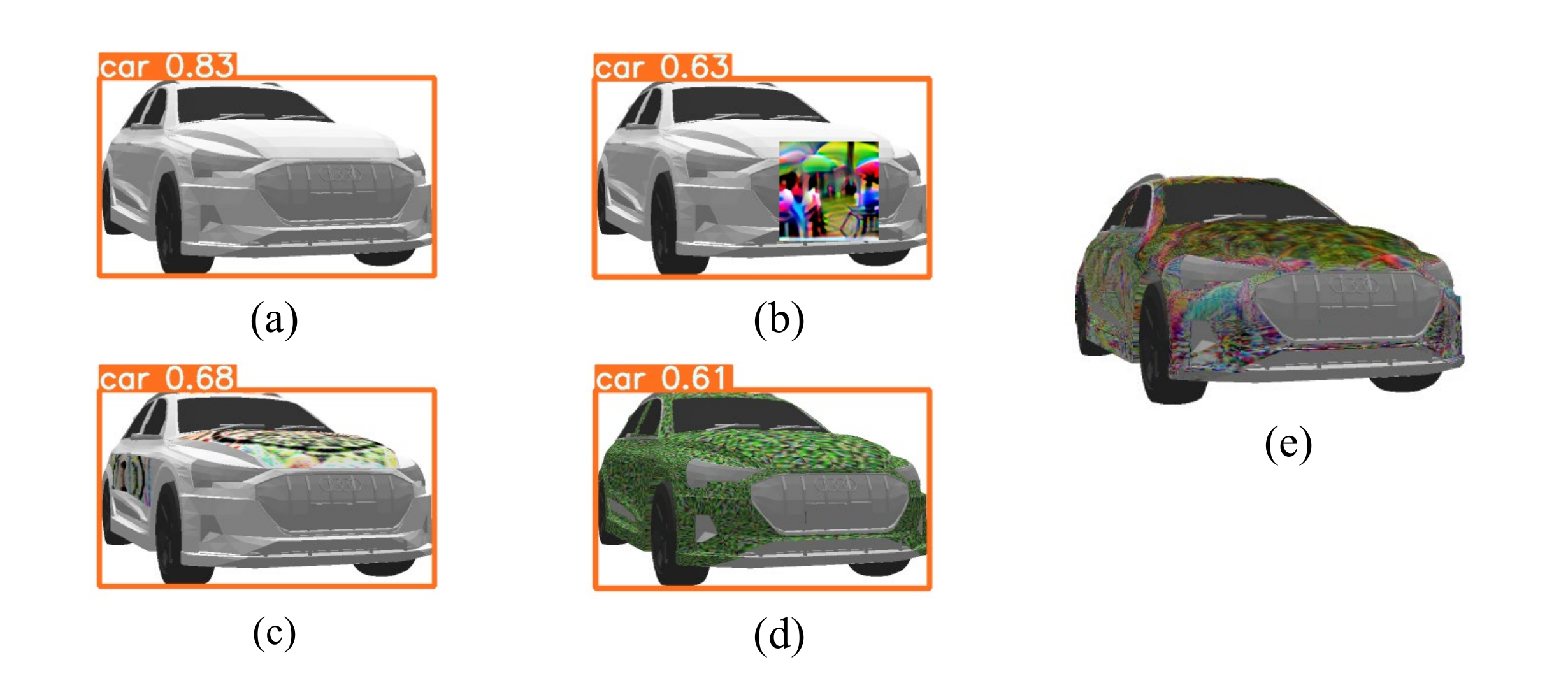}
	\caption{
	(a) is a car without camouflage. 
	(b) is a camouflaged car by placing a planar adversarial patch in front of the car \cite{thys2019fooling}. 
	(c) is a camouflaged car by placing an adversarial patch over the rooftop, hood and doors \cite{wang2021dual}. (d) is a the camouflaged car by repeating an adversarial pattern \cite{zhang2018camou}.
	(e) shows the camouflaged car generated by FCA, which is undetected. }
	\label{fig:example}
\end{figure}

Recently, adversarial attacks on object detection have attracted increasing attention,
particularly in physical attacks due to complex physically realizable constraints (e.g., non-planar object surface) and environmental conditions (e.g., lighting, viewing angles, camera-to-object distances and occlusions)
\cite{elsayed2018adversarial}. 
There are mainly two kinds of methods to modify the visual characteristics of the real object: patch-based and camouflage-based.
\textbf{Patched-based} methods try to perform physical adversarial attacks by generating
adversarial patches\cite{brown2017adversarial}, which confine the noise to a small and localized patch without perturbation constraint. A patch is often stuck to a planar object(e.g., STOP sign \cite{eykholt2018robust}) or placed in front of the object(e.g., person \cite{thys2019fooling}) or placed in the background\cite{lee2019physical}.

\textbf{Camouflage-based} method is implemented by modifying the target object itself and is more challenging due to the non-planarity of 3D objects. 
There are two ways to paint the camouflage: one way is to optimize an adversarial pattern and repeat the patterns as a whole camouflage to paint on the vehicle's surface using a physical non-differentiable  renderer\cite{zhang2018camou, wu2020physical}, while another way is to optimize the texture\cite{zeng2019adversarial,wang2021dual} or the shape\cite{xiao2019meshadv} of the 3D vehicle directly with a differentiable neural renderer.
%

However, existing methods are not robust to specific physical scenarios, especially for multi-view, long-distance and partially occluded objects. 
Firstly, a patch is often stuck to a planar object, so patch-based methods are not suitable and robust for attacking vehicle detectors over 3D vehicles as shown in Figure \ref{fig:example}(b).
Secondly, previous camouflage-based methods \cite{huang2020universal, wang2021dual} paint the adversarial camouflage only on the part of the 3D vehicle model, e.g., the rooftop or side doors, which limits the attack capability in multi-view scenarios when partial adversarial camouflage is not visible, as shown in Figure \ref{fig:example}(c). Besides, \cite{wang2021dual} is not competitive for attacking detectors because they aim to exploit the common characteristic (e.g., model's attention) among models and mainly focus on classifiers.
Thirdly, previous ``full-coverage" camouflaged methods \cite{zhang2018camou, xiao2019meshadv, wu2020physical} generate an individual adversarial pattern and repeat the pattern until covering all the vehicle surface (i.e., as shown in Figure \ref{fig:example}(d)), which is essentially an adversarial image patch optimization. The camouflaged vehicles with image pattern may fail to attack the objectors for multi-view and long-distance scenarios. 

To address the aforementioned problems, we propose an end-to-end Full-coverage Camouflage Attack (FCA) pipeline. Specifically, we first treat the adversarial camouflage as the texture of the 3D vehicle and utilize a neural renderer to paint the texture onto the full surface of vehicle. Then, we apply a transformation function to convert the rendered 3D vehicle into different environment scenarios to get photo-realistic images. And finally, we model the generation of the camouflage texture as an optimization problem by designing an efficient loss function. With such generated adversarial camouflage, the painted vehicle can stay adversarial in physical scenarios for multi-view, long-distance and partially occluded objects.

In summary, our main contributions list as follows.
\begin{itemize}
	\item We bridge the gap between digital attacks and physical attacks via a differentiable neural renderer. We overcome the partial occluded and long-distance issues by painting the adversarial camouflage onto the full vehicle surface.
	\item An end-to-end physical adversarial attack was proposed to generate a robust adversarial camouflage. 
	\item Extensive experiments demonstrate that our method outperforms the existing methods and generalizes to different environments, vehicles, and object detectors.
	In addition, our camouflage can be easily painted or overlaid in the real world and seems natural to humans.
\end{itemize}

\section{Related work}
In this section, we first review the physical adversarial attacks in object detection. And then we briefly introduce the neural renderer.  

\subsection{Physical Adversarial Attack}
According to the implementation methods, the attacks can be briefly divided into patch-based and camouflage-based. 
The patch-based attacks aim to generate an universal image patch\cite{brown2017adversarial}, and several transformations\cite{huang2020universal} were adopted to ensure the transferability. \cite{zhang2018camou} devised a clone network to simulate the process of physical rendered to object predicted, they update the camouflage patch by performing the white-box attack on the clone network. Similarly, \cite{wu2020physical} proposed a query-based discrete searching algorithm to generate an adversarial patch, and then repeated and enlarged the patches until they covered the vehicle surface. Although these attacks achieve certain success, their attacking ability deteriorates when applied to the complex physical world.

The camouflage-based attacks aim to modify the shape or texture of the 3D object. In this category, \cite{xiao2019meshadv} utilized a neural renderer to modify the shape and texture of the textureless object directly, the final result is an adversarial object. Recently, \cite{wang2021dual} proposed a dual attention suppress attack, which suppresses the attention map of the target object in the detection model. To maintain the naturalness of the camouflage (i.e., human attention evasion), they constrain the perturbation only around the content seed. 
In this paper, we paint the texture of the 3D vehicle similarly as \cite{wang2021dual}, however, we find their adversarial camouflage is not robust for multi-view, long-distance and partially occlusions, for which they constrained the camouflage area to the rooftop, hood and car doors. We solve the issue mentioned above with full-coverage (except the glass, tire, lights) camouflage texture.

\subsection{Neural Renderer}
Traditional renderer is commonly used in 2D-to-3D transformation, one of the applications is to wrap the texture image to the 3D model, which then is rendered to the 2D image. To make the rendering process differentiable, \cite{kato2018neural} proposed an approximate gradient for rasterization to enable the integration of rendering into neural networks, which is referred as neural renderer. Initializing  with different camera parameters (i.e., rotation and location), one could render the 3D object model (consisting of mesh and texture) under different view angles. \cite{zhang2018camou} and \cite{wu2020physical} utilized the CARLA\cite{dosovitskiy2017CARLA} simulator to render the adversarial patch onto 3D object, which is non-differentiable. \cite{xiao2019meshadv} used the neural renderer to modify the shape and texture of 3D objects. Following \cite{wang2021dual}, we utilize the neural renderer to paint our adversarial camouflage onto the vehicle surface.

\begin{figure*}[t]
	\begin{minipage}{\linewidth}
		\centering
		\includegraphics[width=0.75\linewidth]{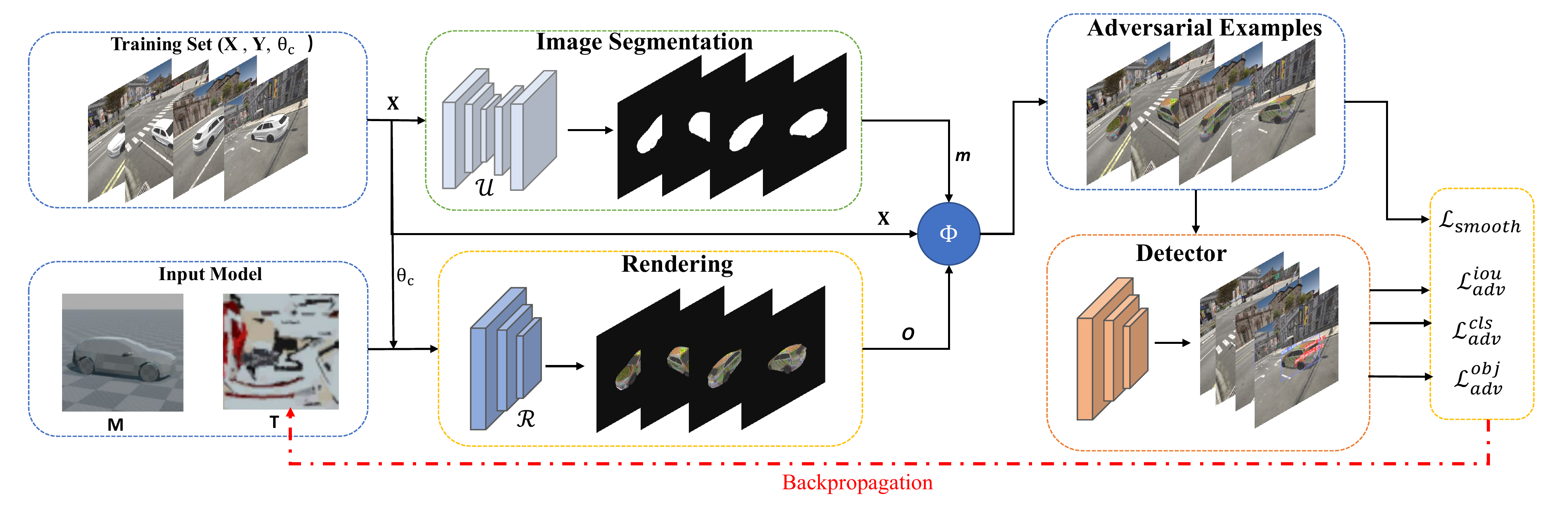}
	\end{minipage}
	\caption{The overview of FCA. Our training set contains the images sampled from the photo-realistic simulator under different simulation settings. We first utilize a pretrained image segmentation network to fetch the target vehicle and binary it as a mask. Meanwhile, we render the camouflage texture onto the surface of the vehicle with the same simulation setting and obtain the camouflaged 2D vehicle. Next, we utilize a transformation function to transfer the camouflaged vehicle into the different physical scenarios with the corresponding mask. Finally, we update the adversarial camouflage through backpropagation with our devised loss function. }
	\label{fig:framework}
\end{figure*}

\section{Method}
In this section, we first introduce the preliminaries. Then we describe the proposed end-to-end physical camouflage adversarial attack in detail.

\subsection{Preliminaries}
Given a vehicle training set $(\textbf{X}, \textbf{Y}, \theta_c)$ where $\textbf{X}$, $\textbf{Y}$ and $\theta_c$ are the sampled images, ground truth labels of the target vehicle and the corresponding camera parameters (i.e., transformation and location) respectively, a 3D vehicle model with a mesh $\mathbf{M}$ and a texture $\textbf{T}$, we use a renderer $\mathcal{R}$ with camera parameter $\theta_c$ to obtain the rendered 2D vehicle image $\textbf{O} = \mathcal{R}(\mathbf{M}, \textbf{T}; \theta_c), \textbf{O} \in \mathbb{R}^{H \times W \times 3}$. To mimic the physical real world, we devise a transformation function $\Phi$ to transfer the rendered vehicle image to different environment scenarios, and then obtain the input image $\textbf{I} \in \mathbb{R}^{H \times W \times 3}$ of the detector. Now, we can obtain the detection result $\textbf{b} = \mathcal{F}(\textbf{I}; \theta_f) = (b_x, b_y, b_w, b_h, b_{obj}, b_{cls}) $, where $\mathcal{F}$ is the object detector with parameters $\theta_f$, $b_x$ and $b_y$ are the center coordinates of the prediction bounding box (i.e., \textit{bbox}), $b_h$ and $b_w$ indicates the height and width of the prediction \textit{bbox}, $b_{obj}$ is confidence score that the \textit{bbox} contains an object, $b_{cls}$ is the class probability distribution of the object in the bbox, \ $b_{cls} \in [1,80]$ in COCO.

Our attack scheme is to generate the adversarial camouflage texture, which can be painted on the surface of the 3D vehicle model. The target vehicle category we select is ``car"  due to the real-time vehicle surveillance is widely used in daily life. Note that our attack target object is only one specific car in the scenario. To realize the adversarial camouflage attack, we replace the origin texture $\mathbf{T}$ with adversarial texture $\mathbf{T}_{adv}$, and obtain the corresponding adversarial image $\mathbf{I}_{adv}$ with transformation function $\Phi$. We aim to hide the target vehicle under the detector $\mathcal{F}$. We treat the adversarial texture generation as an optimization problem, and our objective function is expressed as follows
\begin{equation}
		\mathbf{T}_{adv}^* =  \arg \mathop{\mathrm{max}} \limits_{\mathbf{T}_{adv}} J(\mathcal{F}(\Phi(\mathcal{R}(\mathbf{M}, \mathbf{T}_{adv}; \theta_c)); \theta_f), \textbf{Y})
	\label{eq:object}
\end{equation}

where $\mathbf{T}_{adv}^*$ is the final adversarial texture, $J(\cdot, \cdot)$ is the loss function. By solving the above optimization problem, i.e., Eq \ref{eq:object}, we can obtain the ultimately adversarial camouflage texture.

\subsection{Generating Adversarial Camouflage}

To generate full-coverage adversarial camouflage, we propose an adversarial camouflage texture generation framework with a differentiate neural renderer, which can render the customized texture onto the 3D vehicle model directly. The overall framework of FCA is illustrated in Figure \ref{fig:framework}, our goal is to generate a robust camouflage texture through the backpropagation of loss.

To this end, the loss function plays a vital role in optimizing. In this work, we devise the loss function considering two key aspects: \textit{adversarial loss} to guarantee the attacking ability. \textit{smooth loss} to make the digital-physical difference caused by camouflage more natural. We will discuss these losses in the following sections. 

\subsubsection{Adversarial Loss}
In this work, we use YOLO-V3 as the target detection model $\mathcal{F}$, in other words, we train the adversarial texture with a known model under white-box attack setting. It's well known that YOLO-V3 is a single-stage detector, which makes classify and regression in a single step with dense sampling. Thus it is necessary to take account of attacking both regression and classification simultaneously. After analyzing the loss function of YOLO-V3, we use the following three-loss terms: $\mathcal{L}_{adv}^{iou}, \mathcal{L}_{adv}^{obj}, \mathcal{L}_{adv}^{cls}$. To make the detector incorrectly detected or undetected, we first reduce the intersection over union (IoU) between the prediction \textit{bbox} and ground truth \textit{bbox} to suppress the region of the target prediction \textit{bbox}, which is denoted as $\mathcal{L}_{adv}^{iou}$. Then we reduce the objectness score that indicates whether the prediction \textit{bbox} contains an object by minimizing the objectness confidence. We denote this loss as $\mathcal{L}_{adv}^{obj}$. Finally, to attack the classification, we select the probability of the target object and minimize it, which is denoted as $\mathcal{L}_{adv}^{cls}$. Therefore, our final adversarial loss $\mathcal{L}_{adv}$ is constructed as follow

\begin{equation}
	\mathcal{L}_{adv} = \alpha \mathcal{L}_{adv}^{iou} +  \beta \mathcal{L}_{adv}^{obj} + \gamma \mathcal{L}_{adv}^{cls}
	\label{eq:adv}
\end{equation}

where $\alpha, \beta, \gamma$ are the weights to balance the contribution of each loss term. Then we will exhaustively introduce each loss term of adversarial loss in the following.

$\bullet$ IoU loss $\mathcal{L}_{adv}^{iou}$ represents the overlap area between the ground truth label and the prediction result of the rendered images. One can obtain a high IoU value with a trained detector at the inference stage. By minimizing the $\mathcal{L}_{adv}^{iou}$, we can suppress the prediction \textit{bbox} of the target region. Consequently, the target object is filtered by the detector as the IoU below the threshold. Thus, our $\mathcal{L}_{adv}^{iou}$ is formulated as follows

\begin{equation}
	\mathcal{L}_{adv}^{iou} = \sum_i^N IoU(b^i, b^i_{gt})
\end{equation}

where N denotes the multi-scale (i.e., N=3) output prediction result of the YOLO-V3, $b^i$ and $b^i_{gt}$ is the \textit{i-th} scale prediction result and corresponding ground truth bbox of our attack target, respectively. 

$\bullet$ Objectness loss $\mathcal{L}_{adv}^{obj}$ represents the confidence score whether the detection box contains an object. We follow \cite{thys2019fooling, wang2021towards} and choose the object confidence score to as our $\mathcal{L}_{adv}^{obj}$. 

$\bullet$ Classification loss $\mathcal{L}_{adv}^{cls}$ represents the classification probability of the target class, i.e., car. Specifically, we select the \textit{i-th} scale probability of the target class $t$ in the detection result, denoting it as $b^i_{cls^t}$. Finally, the classification loss can be expressed as

\begin{equation}
	\mathcal{L}_{adv}^{cls} = \sum^N_i b^i_{cls^t}
	\label{eq:cls}
\end{equation}


\subsubsection{Smooth Loss}
To ensure the naturalness of the generated adversarial camouflage, we follow \cite{sharif2016accessorize} to utilize the smooth loss that introduced by \cite{mahendran2015understanding} to reduce the inconsistent among adjacent pixels. For a rendered vehicle image painted with adversarial camouflage $\mathbf{I}_{adv}$, the calculation of smooth loss can be written as
\begin{equation}
	\mathcal{L}_{smooth} = \sum_{i,j}(x_{i,j} - x_{i+1,j})^2  + (x_{i,j} - x_{i,j+1})^2
\end{equation}
 
where $x_{i,j}$ is the pixel value of $\mathbf{I}_{adv}$ at coordinate $(i, j)$.

\subsection{Physical Transformation}
Previous work \cite{wang2021dual} painted the camouflage on the vehicle through \textit{tensor addition}, i.e., the rendered camouflaged vehicle image pixels are directly added to the sampled image containing the original vehicle, which makes it difficult to get convergence during training. Instead, we introduce a simple but efficient approach to substitute the tensor addition. Specifically, we use a segmentation network $\mathcal{U}$ to crop the background from the original photo-realistic image and obtain a binary mask $m\in \mathbb{R}^{H \times W \times 1}$ where the target vehicle areas are set to 1, the background areas are set to 0. With such a mask, we can obtain the adversarial example $\mathbf{I}_{adv}$ by transferring the rendered vehicle image $\mathbf{O}$ into photo-realistic environment scenario. The transformation $\Phi$ can be expressed as follow

\begin{equation}
	\mathbf{I}_{adv} = \Phi(\mathbf{O}) = m \cdot \mathbf{O} + (1 - m) \cdot \mathbf{I}
	\label{eq:transfer}
\end{equation}
where $\cdot$ denotes the pixel-wise multiplication. Note that, we preserve the location and rotation information during the sampling stage of the photo-realistic images, thus the rendered vehicle has the identical orientation as the vehicle in the sampled image.

\subsection{Optimization Process}
Overall, we obtain the adversarial camouflage texture by jointly minimizing the adversarial loss $\mathcal{L}_{adv}$ and smooth loss $\mathcal{L}_{smooth}$. Consequently, our optimization objective can be summarized as 

\begin{equation}
	\mathcal{L}_{total} = \mathcal{L}_{adv} + \mu \mathcal{L}_{smooth}
	\label{eq:total_loss}
\end{equation}

Algorithm \ref{alg:algorithm} summarize the overall training scheme of the presented approach.

\begin{algorithm}[tb]
	\caption{ Full-coverage Camouflage Attack (FCA) }
	\label{alg:algorithm}
	\textbf{Input}: training set $(\textbf{X}, \textbf{Y}, \theta_c)$, 3D model \textbf{(M, T)}, neural renderer $\mathcal{R}$, object detector $\mathcal{F}$, segmentation network $\mathcal{U}$\\
	\textbf{Output}: adversarial texture $\mathbf{T}_{adv}$
	\begin{algorithmic}[1] 
		\STATE Initial $\mathbf{T}_{adv}$ with random noise
		\FOR {the max epochs}
		\STATE select the \textit{minibatch} sample from training set $(\textbf{X}, \textbf{Y}, \theta_c)$
		\STATE  $m$ $\gets$ $\mathcal{U}(\textbf{X})$
		\STATE  $\mathbf{O}$ $\gets$ $\mathcal{R}((\mathbf{M}, \mathbf{T}_{adv}); \theta_c)$
		\STATE  $\mathbf{I}_{adv}$ $\gets$ $m$ $\cdot$ $\mathbf{O}$ + (1 - $m$) $\cdot$ $\mathbf{I}$
		\STATE  $b$ $\gets$ $\mathcal{F}(\mathbf{T}_{adv}; \theta_f)$
		\STATE  calculate $\mathcal{L}$ by Eq \ref{eq:total_loss}
		\STATE  update $\mathbf{T}_{adv}$ with gradient backpropagation
		\ENDFOR
	\end{algorithmic}
\end{algorithm}

\section{Experiments}
In this section, we first describe the experimental settings. 
Then we empirically show the effectiveness
of the proposed full-coverage camouflage by providing thorough evaluations in different simulation environments.

\subsection{Experimental Settings}
\subsubsection{Datasets}
To bridge the gap between digital attacks and physical attacks, we utilize the photo-realistic datasets to perform the experiments. To this end, we select the simulator CARLA \cite{dosovitskiy2017CARLA}, a prevalent open-source  simulator for autonomous driving research, as our 3D simulator. The CARLA simulator provides a variety of high-fidelity digital scenarios (e.g., modern urban) based on Unreal Engine 4. To compare with previous works, we use the same datasets provided by \cite{wang2021dual} directly, the training set consists of 12,500 high-resolution images, while the testing set has 3,000 high-resolution images. The datasets contain images that are sampled from different view angles and distances.

\subsubsection{Evaluation Metrics}
We aim to generate false negatives and hide the target vehicle from the detector. To this end, the first evaluation metric that we select is the Attack Success Rate (ASR) \cite{wu2020making}, which is defined as the percentage of the target vehicles detected before perturbation and not detected or false detected after perturbation. In addition, we adopt the P@0.5 following \cite{zhang2018camou,wang2021dual} as our second evaluation metric, which is defined as the percentage of the correct detected when the detection IoU threshold is set to 0.5. 

\subsubsection{Implementation details}
We choose a widely used detector, YOLO-V3 \cite{redmon2016you}, as our white-box model to train the adversarial camouflage texture. And we evaluate the transferring attack performances (black-box attack) on the following prevalence object detection models: YOLO-V5 \cite{glenn2021yolov5}, SSD \cite{liu2016ssd}, Faster R-CNN \cite{ren2015faster}, and Mask R-CNN \cite{he2017mask}. These models are all pretrained on COCO dataset. Note that, in our experiments, these models are the official implementation version provided by PyTorch \cite{paszke2017automatic} except SSD \footnote{https://github.com/lufficc/SSD}.

The adversarial camouflage texture is initialized as random noise, and the Adam with default parameter is adopted as the optimizer. The hyperparameters are set as follows, the learning rate is 0.01, the max epoch is 5. For $\alpha, \beta, \gamma$, we use the default value 0.05, 1.0, 0.5, respectively, provided by the YOLO-V3 implementation. We follow \cite{wang2021dual} to set the $\mu$ to 1.0. Note that, we also find that hyperparameter $\alpha, \beta, \gamma, \mu$ has a limited impact on our performance in our preliminary experiment. The segmentation network used to extract the background from the photo-realistic image is U2-Net \cite{qin2020u2net}. 
We conduct the experiment on a NVIDIA RTX 3090 24GB GPU cluster.

\subsection{Digital Adversarial Attack}
In this section, we evaluate the performance of adversarial camouflage in the digital space. We report the P@0.5 for the detection of the target vehicle. 

\begin{table}[t]
\caption{The comparison result of adversarial attacks in the digital space.}
\resizebox{0.48\textwidth}{!}{
\begin{tabular}{ccccc}
\hline
	\multirow{2}{*}{Method} & \multicolumn{4}{c}{P@0.5(\%)}                 \\ \cline{2-5} 
	& YOLO-V5 & Faster RCNN & SSD & Mask RCNN \\ \hline
	Raw      & 92.07  &  86.04  &   81.54  &   89.24       \\ 
	MeshAdv   & 72.45  &  71.84  &   66.44  &   80.84        \\ 
	CAMOU    & 74.01  &  69.64  &   73.81  &   76.44       \\ 
	UPC      & 82.41  &  76.94  &   74.58  &   81.97        \\ 
	DAS      & 72.58  &  62.11  &   68.81  &   70.21       \\ 
	DAS-full & 60.52  &  51.43  &   49.93   &  52.07       \\ 
	Ours     & \textbf{32.07}  &  \textbf{34.00}  &   \textbf{28.67}  &   \textbf{30.80}       \\ \hline
\end{tabular}
}
\label{tab:method_compare}
\end{table}

\begin{figure}[htbp]
	\centering
	\includegraphics[width=1\linewidth]{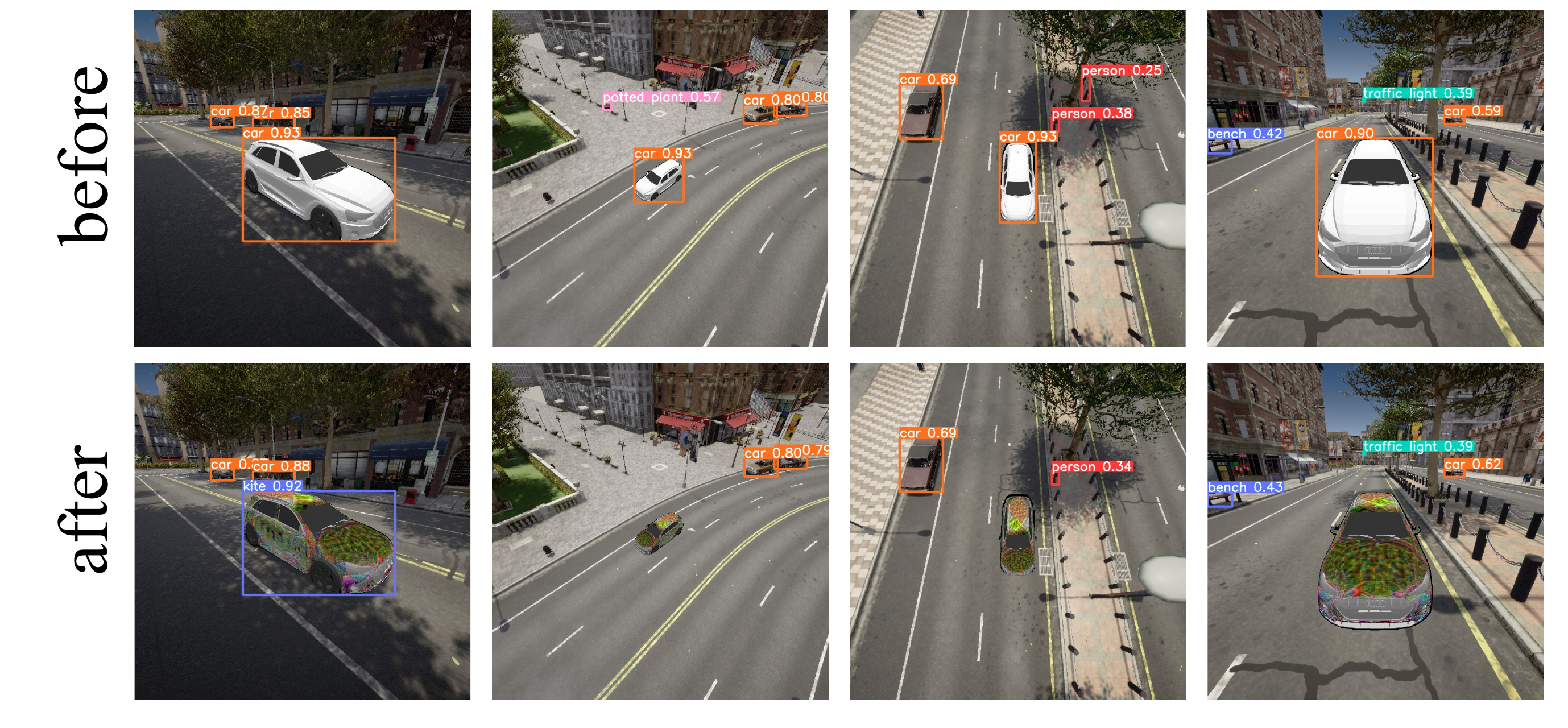}
	\caption{The detection result of the vehicle under different view angles before and after our attack. After painting with our camouflage, the target vehicle turns to be incorrectly detected or undetected.}
	\label{fig:our_camou_case}
\end{figure}

We compare the proposed attacks with several current advanced adversarial camouflage attacks, including MeshAdv \cite{xiao2019meshadv}, CAMOU \cite{zhang2018camou}, UPC \cite{huang2020universal}, DAS \cite{wang2021dual}.
In order to fairly compare our attack with DAS attack, we reimplement the DAS attack with full-coverage camouflage, which denotes as ``DAS-full". 

The comparison results are listed in Table \ref{tab:method_compare}.  Note that, we adopt the results reported in \cite{wang2021dual} because our test set and detectors are identical. As illustrated in Table \ref{tab:method_compare}, our adversarial camouflage significantly outperforms other methods over all the detectors. Specifically, on the one side, the maximum drop of P@0.5 by \textbf{60\%} on YOLO-V5, the minimum drop of P@0.5 by 52.04\% on Faster RCNN, the average drop is 56.02\%, which demonstrate that our attack could successfully paralyze the vehicle detection system. On the other side, in our experiments, Faster RCNN shows better robustness (i.e., lower performance decline) than other baseline detectors, probably due to some modules in Faster RCNN that are robust to the appearance change of the object. Finally, despite DAS use a similar full-coverage camouflage (i.e., DAS-full), our attack still outperforms the DAS, which suggests that our proposed loss function is more suitable for attacking object detection.   

We provide some adversarial camouflage vehicle examples in different scenarios. As illustrated in Figure \ref{fig:our_camou_case}, the vehicle before painted with adversarial camouflage is detected as a car with high detection confidence. However, after painted with our adversarial camouflage texture, the vehicle is detected as other categories, even ``disappear" under the detector. To show the effectiveness of our adversarial camouflage in realizable applications, we provide more diverse examples here\footnote{https://idrl-lab.github.io/Full-coverage-camouflage-adversarial-attack/}. 

\subsection{Physical Adversarial Attack}
In this section, we evaluate the performance of adversarial camouflage in the physical space. We report the P@0.5 for the detection of the target vehicle. 

For simplicity, we compare one partially camouflage attack(i.e., DAS) and two full-coverage camouflage attacks (i.e., CAMOU and DAS-full). Due to the limitation of founds and conditions, we follow \cite{wang2021dual} to print our adversarial camouflages by an HP Color Laser MFP 179fnw printer and crop the camouflage part, then stick them on a toy car with different backgrounds to mimic the real car painting in the physical world. To show the efficiency of our adversarial camouflage under different scenarios, we capture 144 pictures of the painted car on different settings (i.e., 8 directions ( $45^\circ$ / $360^\circ$), 3 distances {long, middle, and short distance}, 3 different surroundings) with a Redmi K20 Pro phone.


\begin{table}[t]
\caption{The comparison result of adversarial attacks in the physical space.}
\resizebox{0.48\textwidth}{!}{
\begin{tabular}{ccccc}
\hline
	\multirow{2}{*}{Method} & \multicolumn{4}{c}{P@0.5(\%)}                 \\ \cline{2-5} 
	& YOLO-V5 & Faster RCNN & SSD & Mask RCNN \\ \hline
	Raw      & 100  &  88.89    &   78.47  &   96.53       \\ 
	CAMOU    & 72.22  &  44.44  &   40.97  &   53.48       \\ 
	DAS      & 100  &  65.28    &   52.08  &   68.06       \\ 
	DAS-full & 94.44  &  43.06  &   43.75   &  45.83       \\ 
	Ours     & \textbf{65.28}   &  \textbf{24.31}  &   \textbf{29.17}  &   \textbf{29.17}       \\ \hline
\end{tabular}
}
\label{tab:phy_compare}
\end{table}

The evaluation results are list in Table \ref{tab:phy_compare}. Compared with other methods, the FCA can transfer to the physical world well, we get 65.28\% on YOLO-V5, 24.31\% on SSD, 29.17\% on Faster RCNN, 29.17\% on Mask RCNN, respectively. That indicates that FCA can pose more potential risks for the detection systems in the real world. Moreover, all the full-coverage adversarial camouflages are better than the partial coverage adversarial camouflage, which is consistent with our analysis that the performance of existing adversarial camouflage attacks degrades due to multi-view or partially occlusion scenarios. However, the Mask RCNN shows the worst robustness (maximum drop in P@0.5). While the YOLO-V5 shows the best robustness against adversarial camouflage, which may be attributed to the special design that makes it more suitable for real-world application. Despite this strong model, our FCA method can also degrade the detection performance in a large marginal, which demonstrated that our adversarial camouflage has a strong transferable attacking ability in the physical world.

\subsection{Multi-view Robust Attack}
\subsubsection{Robustness of Long Distance}

To demonstrate the robustness of our adversarial camouflage in multi-view and long-distance scenarios, we conduct extensive experiments. Specifically, in the experiment, the camera distance we used includes [1.5, 3, 5, 10, 15, 20], the camera elevation we used includes [0, 10, 20, 30, 40, 50] (0 indicates that the camera and the vehicle are parallel). We sample an image every $3^\circ$ a time in $360^\circ$. For a fixed combination of camera distance and elevation, we obtain 120 images, and collect 4320 test images in total. To better illustrate the result, we regroup the rendered image test set in terms of azimuth ranges (i.e., every $45^\circ$ azimuth) and camera distances, then every group has 90 test samples with various camera elevation, in other words, every item in Table \ref{tab: multiview} is conducted on different 90 test images. Note that to better evaluate the view angles and distances without considering different background environments, we use the rendered images $\mathbf{O}$ (pure background) directly for simple implementation. We use YOLO-V5 to evaluate the test images as other detectors exhibit similar trends.
	
The results are listed in Table \ref{tab: multiview}. We can observe that we achieve 100\% ASR in a majority of cases where the distance is among 1.5, 3 and 20. Meanwhile, we find that along with the distance increase, the ASR first prone to decrease at the distance of 10, after that the trend of ASR prone to increase. By contrast, the images sampled at distance of 5, 10 and 15 are hard to attack, which demonstrates the detector is more robust for such settings. Nevertheless, our adversarial camouflage can achieve nearly perfect performance without retrained on the rendered images $\mathbf{O}$, which demonstrate that the generated adversarial camouflage has well transferability across different domain datasets.


\begin{table}[]
	\centering
	\small
	\caption{The ASR (\%) performance for multi-view and multi-distance attack.}
	\begin{tabular}{ccccccc}
		\hline
		\multirow{2}{*}{Azimuth $(^\circ$)} & \multicolumn{6}{c}{Distance} \\ \cline{2-7} 
						& 1.5    & 3          & 5            &  10      &  15      &  20     \\ \hline
		$0 \sim 45$    & 100     & 100         &   84.27      &  68.6    &   80.85  &  100   \\ 
		$ 45 \sim 90$  & 95.83   &  93.33      &   88.89      &  81.82   &   90.2   &  100   \\ 
		$ 90 \sim 135$ & 100     & 100         &   88.31      &  87.5    &   94.92  &  100   \\ 
		$135 \sim 180$ & 100     & 100         &   84.44      &  71.11   &   78.57  &  87.5   \\ 
		$180 \sim 225$ & 100     & 100         &   95.51      &  92.22   &   88.24  &  100   \\ 
		$225 \sim 270$ & 100     & 100         &   98.65      &  88.57   &   95.65  &  100   \\ 
		$270 \sim 315$ & 100     & 100         &   92.96      &  86.96   &   95.65  &  100   \\ 
		$315 \sim 360$ & 100     & 100         &   94.44      &  74.44   &   83.33  &  100   \\ \hline
	\end{tabular}
	\label{tab: multiview}
\end{table}

\subsubsection{Robustness of Partial Occlusion}
\begin{table}[]
	\centering
	\caption{The ASR performance for partially occluded objects for different distances.}
	\resizebox{0.4\textwidth}{!}{
	\begin{tabular}{ccccc}
		\hline
		\multirow{2}{*}{Occlusion} & \multicolumn{4}{c}{Distance} \\ \cline{2-5} 
		       & 1.5    & 3       & 5        &   10          \\ \hline
		small  &   100  &  100    &  62.22   &   62.68   \\ 
		middle &   98   &   92.05 &  78.89   &   77.14    \\ 
		large  &   96   &  97.62  &  72.86   &   78.57  \\ \hline
	\end{tabular}
	}
	\label{tab: occluded}
\end{table}
We also investigate the robustness of our attack when the adversarial camouflage is partially occluded. According to the area of the occluded camouflage, we group the partial occlusion into small occlusion, middle occlusion and large occlusion. Specifically, we define the large, middle and small partial occlusion as the $ \geq 70\%, 30\% \sim 70\%,  \leq 30\%$ of car body is occluded, respectively. In this experiment, We only use the [1.5, 3, 5, 10] camera distances due to the occluded rendered object is too small when the camera distance exceeds 10. For each group and a given camera distance, we collect 90 test images, and totally collect 1080 test images .We use the YOLO-V5 as our evaluation model. 

\begin{figure}[htbp]
	\centering
	\includegraphics[width=1\linewidth]{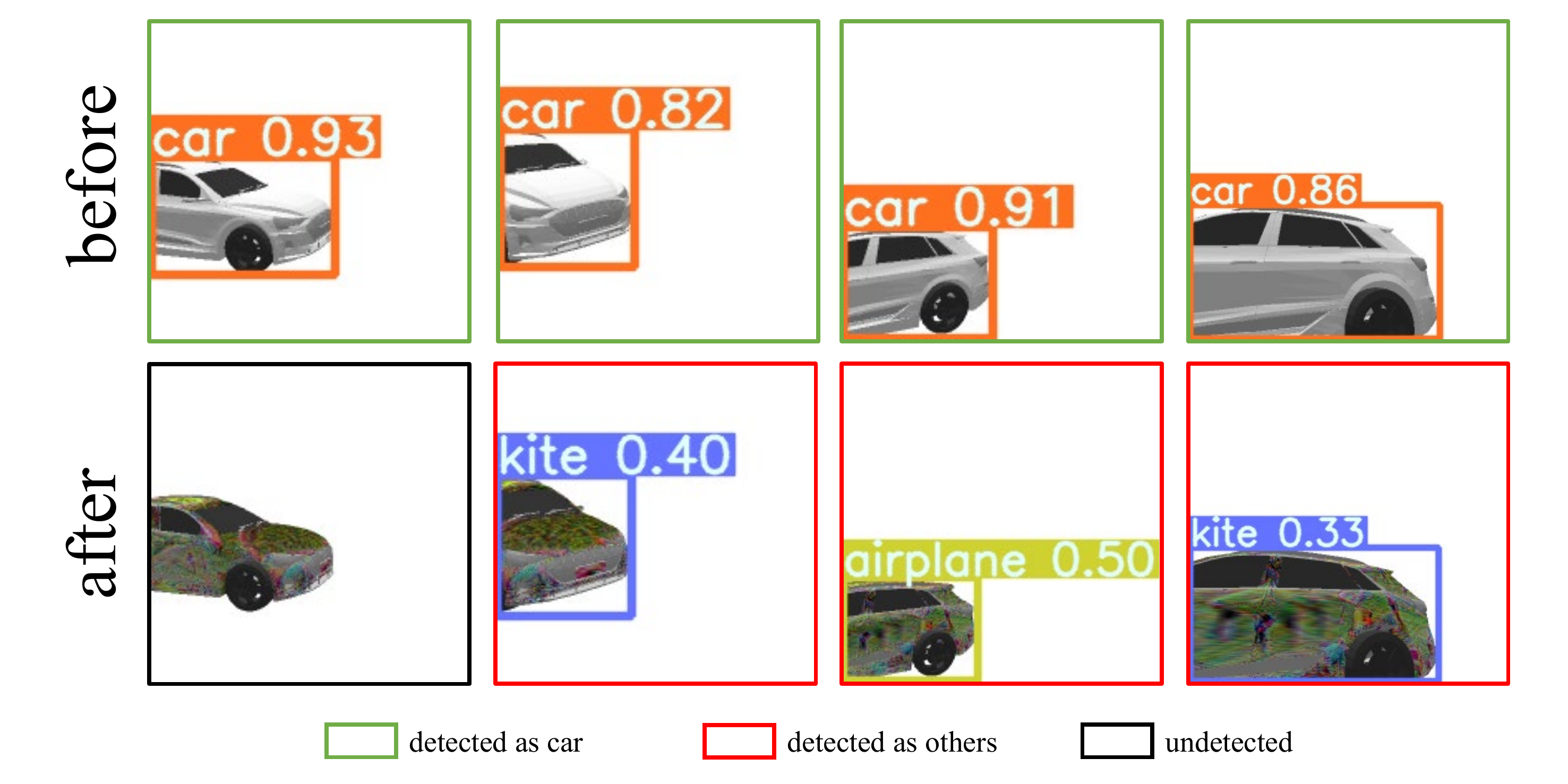}
	\caption{The case of occlusion vehicle before and after our attack. After painted with our camouflage, the detector incorrectly detected or not detected the ``car".}
	\label{fig:case_study}
\end{figure}

Results are listed in Table \ref{tab: occluded}. As we can see the generated camouflage works well at 1.5 and 3 camera distances, particularly in a small occlusion (ASR achieve 100\%), which is attributed to the ratio of rendered images as well as the camouflages are relatively large. On the other hand, when the camera distance exceeds 5, the performance degrades sharply, the possible reason is that the rendered object in images is a very small object and the camouflage being occluded further leads to performance decreasing. We provide some partial occlusion cases in Figure \ref{fig:case_study}, which demonstrated our adversarial camouflage works well for most partial occlusion scenarios. Our camouflage is more robust for occlusion when the camera distance is less than 3, while the robustness trends to degrade when the camera distance increases. In conclusion, our generated adversarial camouflage is robust to different levels of partial occlusions. 

\subsection{Enhance the Transferability}
It has been proved that transferability of adversarial examples can benefit from hard examples \cite{liu2020bias, xie2019improving}. Thus, we argue that the transferability of our camouflage can be further enhanced with hard examples. To this end, we define the attack failure examples as hard examples, then we collect them across the different detectors, and utilize the YOLO-V3 to fine-tune the adversarial camouflage texture on hard examples. Note that, the ratio of hard examples for YOLO-V5, Faster RCNN, Mask-RCNN, SSD is $17.1\%, 29.43\%, 27.71\%, 17.38\%$, respectively. And the number of the enhanced dataset is 4932. 

The updated results are listed in Table \ref{tab:enhanced}, the row indicates the detector that used to generate failure examples and the column indicates the re-evaluated results with fine-tuned hard examples. The diagonal entries of the table indicate the detector used to collect failure examples and re-evaluate is identical. From the table, we can observe that hard examples can enhance the transferability of the adversarial camouflage, we obtain 2.43\% gain for YOLO-V5 itself. However, we also notice that the hard examples are not always effective, the ASR on Faster RCNN and Mask RCNN of all fine-tuning adversarial textures even degrades compared to unenhanced results. To explain this phenomenon, we analysis the collected hard examples, and find the union of these hard examples achieves 41.1\% of the training set, while the intersection of these hard examples is nearly 0\%. The reason may attribute to the different architecture of the detector, some failure examples collected by four detectors may still successfully attack YOLO-V3 during fine-tuning, which means such hard examples are helpless for improving transferability.

\begin{table}[t]
	\caption{The ASR on four detectors where we retrain the camouflage texture on different hard examples extracted by various detectors. The diagonal entries indicate retrained and evaluated on the same detector, while the off-diagonal entries indicate the transfer attack.}
	\small 
	\centering
	\resizebox{0.48\textwidth}{!}{
	\begin{tabular}{ccccc}
		\hline
		\multirow{2}{*}{Method} & \multicolumn{4}{c}{ASR(\%)}                 \\ \cline{2-5} 
						& YOLO-V5 & Faster RCNN & SSD & Mask RCNN \\ \hline
		unenhanced             & 87.67          &    72.11  &   78.42        &  75.16       \\ \hline
		YOLO-V5 		& \textbf{90.1}  &    71.35  & \textbf{79.17} &  73.99       \\ 
		Faster RCNN     & \textbf{88.92} &    70.8   & \textbf{79.65} &  74.06         \\ 
		SSD		        & 86.97          &    70.04  &  78.06         &  73.17       \\ 
		Mask RCNN        & \textbf{89.8} &    70.44  & \textbf{79.43} &  74.13         \\ \hline
	\end{tabular}
	}
	\label{tab:enhanced}
\end{table}

\subsection{Interpretability of the Adversarial Camouflage}
\begin{figure}[htbp]
	\centering
	\includegraphics[width=1\linewidth]{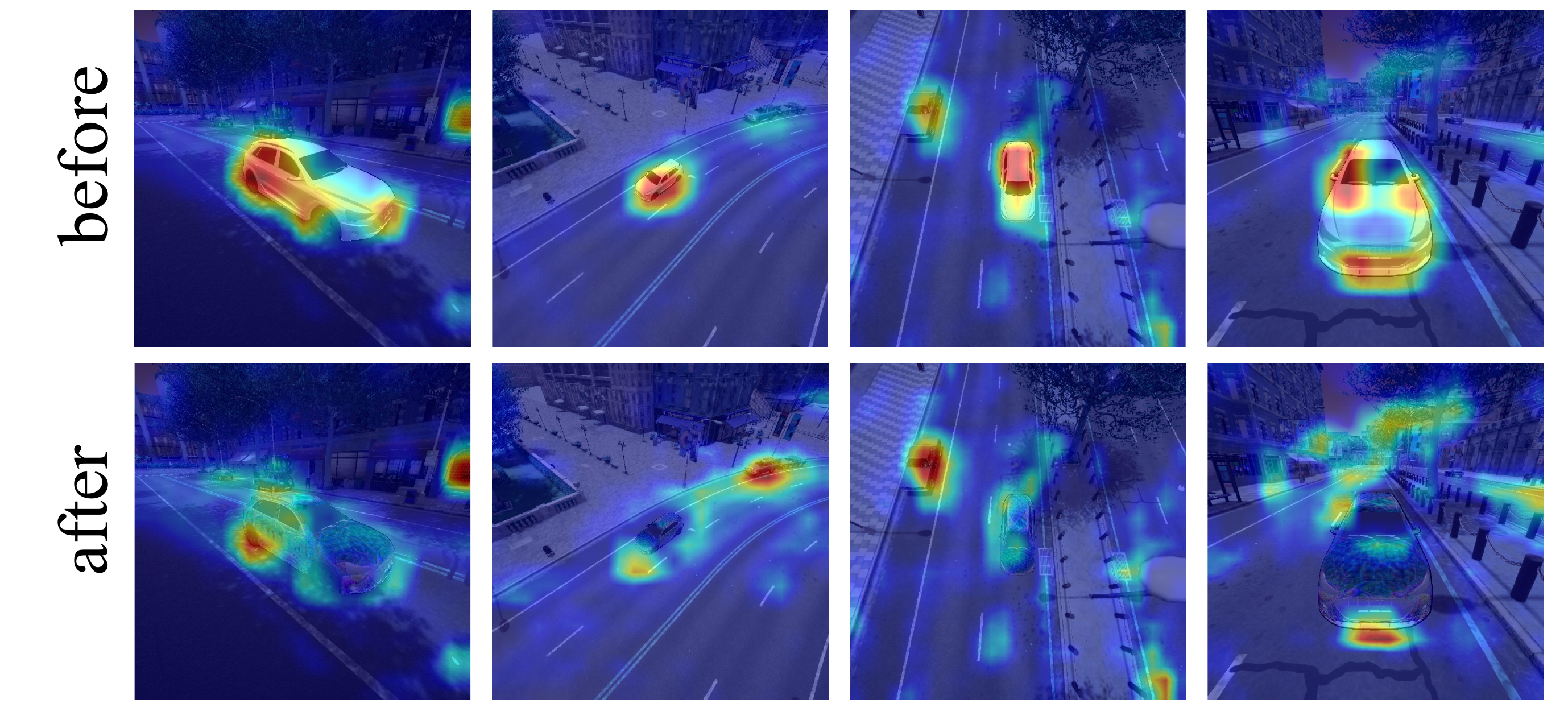}
	\caption{The attention map of the vehicle before and after our attack. After painting our full-coverage camouflage, the attention of the vehicle is dispersed in the image.}
	\label{fig:cam}
\end{figure}

In this section, we try to explain why the detector fails on our generated adversarial camouflage. Following \cite{wang2021dual}, we choose the commonly used interpretability technique, i.e. Grad-CAM \cite{selvaraju2017grad}. We use ResNet50\cite{he2016deep} that pretrained on ImageNet as the base model to extract the attention map of the target vehicle category, the results are illustrated in Figure \ref{fig:cam}. We can see the model's attention on the target category is dispersed after painting our camouflage, which suggested that the decision evidence of the model has been changed. Therefore, the detector makes incorrect inference on adversarial examples.

\subsection{Ablation Studies}
In this section, we investigate the influence of the loss function items and the initialization ways of the camouflage texture.

\subsubsection{Effectiveness of the combination of loss terms.}
Different loss items have different effects. In this part, we conduct the following two-fold studies: the first fold compares $\mathcal{L}^{cls}_{adv}$, $\mathcal{L}^{obj}_{adv}$, $\mathcal{L}^{iou}_{adv}$, $\mathcal{L}^{cls+obj}_{adv}$, which all contain the smooth loss. The second fold investigates the influence of smooth loss in our method, we denote the loss containing only adversarial loss as $\mathcal{L}_{adv}$.  $\mathcal{L}_{total}$ denotes a combination of both adversarial loss and smooth loss. We optimize the adversarial camouflage with different loss term schemes, and evaluate the ASR performance on different detectors. The experiment results are shown in Table \ref{tab:loss_compare}.

\begin{table}[t]
	\caption{The comparison results of different loss schemes.}
	\small 
	\begin{tabular}{ccccc}
	\hline
	\multirow{2}{*}{Method} & \multicolumn{4}{c}{ASR (\%)}                 \\ \cline{2-5} 
	& YOLO-V5 & Faster RCNN & SSD & Mask RCNN \\ \hline
	$\mathcal{L}^{cls}_{adv}$                     & 82.83 &   65.25  &  73.20   &   67.97        \\ 
	$\mathcal{L}^{obj}_{adv}$                     & 46.16 &   49.69  &  61.87   &   42.83         \\ 
	$\mathcal{L}^{iou}_{adv}$                     & \textbf{90.96} &   55.84  &  71.19   &   54.87        \\ 
	$\mathcal{L}^{cls+obj}_{adv}$          & 84.49 &   68.16  &  76.05   &   71.99        \\ \hline
	$\mathcal{L}_{adv}$           	   		 & 88.59 &   \textbf{73.54}  &  \textbf{79.38}   &   \textbf{75.60}        \\ 
	$\mathcal{L}_{total}$				             & 87.67 &   72.11  &  78.42   &   75.16        \\ \hline
	\end{tabular}
	\label{tab:loss_compare}
\end{table}

As we can observe from Table \ref{tab:loss_compare}, on the first fold, we obtain the highest ASR in YOLO-V5 with $\mathcal{L}^{iou}_{adv}$, exceeding 90\%. On the contrary, the ASR of other three models is relatively low. We conclude that the $\mathcal{L}^{iou}_{adv}$ has significant impact on the ASR in particular detectors. On the second fold, the ASR without smooth loss is higher than that with smooth loss for all models, while smooth loss makes the camouflage more natural to humans. In summary, the devised $\mathcal{L}_{total}$ balances the attack performance and the naturalness of the adversarial camouflage, and the $\mathcal{L}^{cls}_{adv}$ and $\mathcal{L}^{iou}_{adv}$ make considerable contributions to the attack.

\begin{table}[t]
	\caption{The comparison result of different texture initialization.}
		\small 
	\begin{tabular}{ccccc}
		\hline
		\multirow{2}{*}{Method} & \multicolumn{4}{c}{ASR  (\%)}                 \\ \cline{2-5} 
		& YOLO-V5 & Faster RCNN & SSD & Mask RCNN \\ \hline
		basic                    & 84.56     &    68.41   &  \textbf{80.65}   &     69.44      \\ 
		random                  & 87.67 &       72.11     &  78.42   &      75.16     \\ 
		zero                    & \textbf{89.9} &        \textbf{74.37}    &   79.79  &       \textbf{75.81}    \\ \hline
	\end{tabular}
	\label{tab:init_compare}
\end{table}
\subsubsection{Effectiveness of different initialization.}
Initialization plays an important role in deep learning, we investigate the influence on the initialization of adversarial camouflage in this part. We mainly compare three different initialization ways: the original basic texture of the 3D model, random noise and zero. As shown in Table \ref{tab:init_compare}, we can observe that the performance of the zero initialization is superior over the other two ways, giving the highest ASR 89.9\% over YOLO-V5, the performance of the original 3D model texture is worse than other two ways on attack Faster RCNN and Mask RCNN, which is less than 70\% (68.41\% for Faster RCNN, 69.44\% for Mask RCNN). This phenomenon may be attributed to that we adopt the gradient descent algorithm to guide the adversarial camouflage update, the random noise initialization gives a prior knowledge that may directly mislead the detector, resulting in wrong optimization directions. In conclusion, the initialization has limit influence on the attack performance, and thus we select the random initialization for balancing the attack and naturalness.

\section{Conclusion}
In this paper, we propose an end-to-end attack method to generate a full-coverage adversarial camouflage in the physical world. Specifically, we first utilize a neural renderer to render our camouflage texture into a 3D vehicle model. Then we devise a transformation function to transfer the rendered vehicle into the photo-realistic simulation scenarios to simulate the complex real-world environmental conditions. Finally, we devise an adversarial loss functions to guide the optimization of camouflage with a gradient descent algorithm. Extensive experiments demonstrated that our FCA outperforms other advanced attacks, and achieves higher attack performance on both digital and physical attacks. Therefore, our method can bridge the gap between digital attacks and physical attacks as much as possible. We hope the proposed FCA could provide an interesting direction of physical attack for future work.

\section{Acknowledgments}
This work was supported in part by National Natural Science Foundation of China under Grant No.11725211, 52005505, and 62001502, and Post-graduate Scientific Research Innovation of Hunan Province under Grant No.CX20200006.

\bibliography{references}

\end{document}